\documentclass[letterpaper]{article}
\usepackage{aaai2027}
\nocopyright
\usepackage[hyphens]{url}
\usepackage{graphicx}
\usepackage{natbib}
\usepackage{caption}
\usepackage{amsmath}
\usepackage{amssymb}
\usepackage{booktabs}
\usepackage{multirow}
\usepackage{xcolor}

\newcommand{\tinypm}[1]{\,{\scriptsize$\pm$#1}}

\title{Capacity, Not Format:\\Rethinking Structured Reasoning Failures}

\author{
    Hengxin Fan
}
\affiliations{
    Tianjin Normal University
}

\begin{document}
\maketitle

\begin{abstract}
Prior work treats structured output as a reasoning tax, but this framing is incomplete: the cost of formatting depends strongly on a model's spare capacity. Using information-matched prose controls and a four-level schema complexity gradient, we separate format-specific effects from prompt-length confounds across 4 models and 5 benchmarks with 0\% parse failures on successfully generated responses. 

We find that structured formats are \emph{capacity-dependent}. Models with sufficient headroom absorb JSON constraints without degradation (Sonnet: $88.7\pm4.0$\% JSON vs.\ $89.3\pm1.7$\% CoT on MATH-Hard). In contrast, formats severely degrade models operating near their limits through two distinct mechanisms. First, under standard token budgets, Haiku drops 36.2pp ($p < 0.0001$) largely due to truncation. Second, even with extended budgets eliminating truncation, GPT-4o-mini drops 28.0pp ($p < 0.001$), revealing pure capacity competition independent of token exhaustion.

This format penalty scales with schema complexity (McNemar $p < 0.0001$) and cannot be explained by prompt length alone. Furthermore, these results qualify claims of frontier model immunity: on AIME competition math, Opus~4.7 drops from 96.2\% to 91.0\% under JSON ($-$5.3pp\footnote{The displayed percentages are independently rounded; the exact difference is $7/133 = 5.26$pp $\approx 5.3$pp.}). A delayed-structure ablation---reasoning freely before formatting---recovers most of the lost accuracy (3-run mean: 80--87\%), supporting the capacity competition mechanism. The practical implication is not to avoid structured output, but to match it to capacity: when a model is near its limits, think first, format later.
\end{abstract}
\section{Introduction}
\label{sec:intro}

When does structured output hurt LLM reasoning? Recent work frames structured formats as a ``tax'' on reasoning \citep{tam2024let,constrainttax2025,snowballing2026,lee2026format}. But this framing is incomplete: it treats format effects as uniformly negative, varying only in magnitude. We show that the same JSON schema is absorbed without cost by capable models while severely degrading weaker ones---on the same task.

The key variable is not the format but the model's \textbf{spare capacity}. Structured output imposes no penalty when a model has headroom beyond the task's demands: under 3-run replication, Sonnet~4.6 scores identically under JSON, free-form prose, and chain-of-thought on MATH-Hard ($\sim$89\%). But when the task pushes a model near its capability boundary, format compliance competes with reasoning for the same limited generation capacity. The apparent immunity of frontier models is not intrinsic; it reflects evaluation on tasks within their comfort zone.

We study 4 models spanning 2 provider families and 2 capability tiers (Claude Sonnet~4.6, Claude Haiku~4.5, GPT-4o, GPT-4o-mini), 5 benchmarks covering math, logic, and knowledge reasoning (GSM8K, MATH-Hard Level~5, BBH, MMLU-Pro), and 4 output conditions including a novel information-matched prose control. Our contributions:

\begin{enumerate}
    \item \textbf{Capacity-dependent format effects.} Using information-matched prose controls and a four-level schema gradient, we show that structured output is neutral for models with spare capacity (Sonnet: $\sim$89\% across all conditions under 3-run replication) while severely degrading capacity-limited models (Haiku: $-$36pp). A unified item-level logistic regression confirms this difficulty $\times$ capability interaction ($p = 3.8 \times 10^{-3}$ under clustered standard errors). Frontier probes on AIME and GPQA Diamond are directionally consistent with the model's predictions.

    \item \textbf{Controlled isolation of format-specific effects.} A detailed prose prompt---information-matched to the JSON instruction---produced 80.7\% accuracy for Haiku vs.\ 52.5\% with JSON ($p < 0.0001$), with 0\% parse failures across all conditions. The degradation is attributable to simultaneous reasoning and schema-compliant serialization, not prompt length or extraction artifacts.

    \item \textbf{Evidence that frontier closed-weight models are not immune.} On AIME competition math, Opus~4.7 drops from 96.2\% to 91.0\% under JSON ($-$5.3pp), qualifying claims that modern closed models show little or no format tax.

    \item \textbf{Mechanistic evidence from failures and interventions.} An 81-case failure analysis reveals that error profiles shift with capability (reasoning omission for weak models, arithmetic errors for stronger ones). A delayed-structure ablation recovers 80--87\% of lost accuracy for capacity-limited models (GPT-4o consistently exceeds CoT under delayed prompting, suggesting additional scaffolding benefit). A token-budget ablation shows the mechanism differs by model: Haiku's tax is largely driven by token exhaustion, while GPT-4o-mini's penalty ($-$28pp) persists at 0\% truncation even with 4$\times$ the token budget, revealing pure capacity competition independent of output length.
\end{enumerate}

These findings reframe the ``format tax'' narrative: structured output is not inherently harmful---it is a capacity demand that strong models absorb and weaker models struggle with. The practical remedy is not to avoid structure, but to match it to capacity: \emph{think first, format later}.

\section{Related Work}
\label{sec:related}

\paragraph{Structured Output and Reasoning.}
\citet{tam2024let} evaluated JSON, XML, and YAML format restrictions across multiple models and found that format-restricting instructions caused significant reasoning degradation, with stricter constraints producing larger declines. \citet{long2025biased} showed that LLMs exhibit substantial output format bias, with performance varying significantly across formats. \citet{constrainttax2025} quantified the validity-correctness tradeoff: hard schema enforcement achieved 100\% format validity but reduced answer accuracy by up to 8.7 percentage points. \citet{snowballing2026} identified ``structure snowballing,'' where models prioritize format compliance over reasoning during reflective generation. In grammar-constrained generation, CRANE \citep{crane2025} showed that augmenting restrictive grammars with reasoning-permitting rules can preserve reasoning while maintaining syntactic and semantic correctness. Concurrent with our work, \citet{lee2026format} systematically benchmark the ``format tax'' across models and tasks, finding that structured output degrades reasoning---particularly for open-weight models---and that decoupling reasoning from formatting recovers most of the loss. They conclude that most closed-weight models show little to no format tax. Our work differs in two key respects: (i) we show that even frontier closed-weight models (Opus~4.7) exhibit degradation when evaluated on tasks near their capability boundary (AIME: $-$5.3pp), suggesting their apparent resilience reflects insufficiently challenging evaluation rather than genuine immunity; and (ii) we introduce an information-matched prose control that isolates format-specific effects from prompt complexity, ruling out the confound that format-requesting instructions are simply longer or more detailed than free-form prompts.

\paragraph{Causal Analysis of Format Effects.}
\citet{yuan2025causal} applied causal inference and found no causal impact of output format in 43 of 48 scenarios, suggesting previously reported effects may be confounded. Our GSM8K results ($\pm$4\% effects) are consistent with their null findings. We go further by explaining \emph{when} significant effects emerge---on tasks that stress model capacity---and by introducing an information-matched prose control that isolates format-specific effects from prompt complexity, a confound not isolated in prior work using an information-matched prose control.

\paragraph{Chain-of-Thought and Structured Reasoning.}
Chain-of-thought prompting \citep{wei2022chain} established that the structure of generation affects reasoning quality. Subsequent work explored richer structures: Tree of Thoughts \citep{yao2023tree} enables deliberate exploration via tree-structured reasoning, Self-Discover \citep{zhou2024self} composes task-specific reasoning structures from atomic modules, and ReasonFlux \citep{reasonflux2025} scales hierarchical thought templates. These approaches show that semantically meaningful structure enhances reasoning---consistent with our observation that strong models are not harmed by JSON's step-by-step scaffold when they have sufficient spare capacity. The distinction we identify is that this tolerance requires sufficient headroom: models near their limits struggle to afford the overhead.

\paragraph{Scaling and Capability Effects.}
Scaling laws \citep{kaplan2020scaling} established that capabilities vary predictably with scale, and \citet{wei2022emergent} showed some capabilities emerge abruptly. Our finding adds a new dimension: the same prompting technique can be neutral for strong models but severely harmful for weaker ones, implying format choice should be capability-aware.

\section{Experimental Setup}
\label{sec:setup}

\subsection{Models}
We evaluate 4 models spanning 2 provider families and 2 capability tiers:

\begin{itemize}
    \item \textbf{Strong tier:} Claude Sonnet~4.6 (\texttt{claude-sonnet-4-6}) and GPT-4o (\texttt{gpt-4o}).
    \item \textbf{Weak tier:} Claude Haiku~4.5 (\texttt{claude-haiku-4-5}) and GPT-4o-mini (\texttt{gpt-4o-mini}).
\end{itemize}

We operationally define model capability as baseline chain-of-thought accuracy on MATH-Hard (our hard benchmark). Under this definition, Sonnet (89.3\%) and Haiku (88.7\%) have similar 3-run mean baselines on MATH-Hard, yet diverge dramatically under format constraints---suggesting that raw accuracy alone does not predict format resilience. Our model set intentionally spans deployed mid-capacity systems (GPT-4o, GPT-4o-mini) and contemporary frontier systems (Claude Sonnet~4.6, Haiku~4.5, Opus~4.7). We do not interpret release year as the causal factor; rather, we analyze format tax as a function of observed model capability.

\subsection{Benchmarks}
We use five benchmarks spanning math, logic, and knowledge reasoning:

\paragraph{GSM8K} \citep{cobbe2021training}. Grade-school math word problems. We randomly sample 100 problems (seed=42) from the test set. All 4 models achieve $\geq$90\% accuracy under chain-of-thought, establishing that this benchmark lies within their comfort zone.

\paragraph{MATH-Hard (Level 5)} \citep{hendrycks2021measuring}. Competition-level mathematics from the MATH dataset, filtered to Level~5 (hardest). We sample 100 problems with clean numerical answers. Baseline accuracies range from 56.2\% (GPT-4o) to 89.3\% (Sonnet), placing several models near their capability boundaries.

\paragraph{BBH Logical Deduction} \citep{suzgun2023challenging}. From BIG-Bench Hard: 7-object logical deduction requiring multi-step constraint satisfaction. 100 problems, multiple-choice (A--G). Baselines range from 79.0\% (GPT-4o-mini) to 97.0\% (Haiku).

\paragraph{BBH Tracking Shuffled Objects} \citep{suzgun2023challenging}. State tracking through a sequence of swaps among 7 objects. 100 problems. Most models achieve $\geq$98\%, but GPT-4o-mini scores 90.0\%, providing a capability gap.

\paragraph{MMLU-Pro} \citep{wang2024mmlu}. Graduate-level 10-way multiple choice across law, physics, and philosophy (100 problems each, 300 total). Baselines range from 51.3\% (GPT-4o-mini) to 83.3\% (Sonnet), making this our hardest non-math benchmark.

\subsection{Output Format Conditions}
\label{sec:conditions}

We test 4 output conditions, designed to isolate the effect of structured formatting from prompt complexity:

\paragraph{Chain-of-thought (CoT).} A minimal prompt: ``Solve the math problem. Think step by step, then give your final numerical answer.'' (82 characters.)

\paragraph{JSON-structured (JSON).} A detailed JSON schema requiring the model to produce a structured response with fields for problem understanding, known values, approach, step-by-step solution with verification, and final answer. (1,774 characters.)

\paragraph{XML-structured (XML).} An equivalent schema in XML format, requiring the same fields and level of detail. (1,625 characters.)

\paragraph{Detailed prose (DP).} A information-matched natural language prompt requesting the same analytical dimensions (problem understanding, known values, approach, step-by-step with verification) but without any format constraint---the model responds in free-form text. (1,280 characters.)

The detailed prose condition is our key methodological contribution. By matching the informational content and approximate length of the structured format prompts without imposing any syntactic constraints, it allows us to decompose observed effects into prompt complexity (DP vs.\ CoT) and format-specific (JSON/XML vs.\ DP) components.

\subsection{Delayed Structure Ablation}
\label{sec:delayed}

To test whether simultaneous formatting and reasoning compete for model capacity, we introduce a \emph{delayed structure} condition: the model first reasons freely in natural language (Phase~1), then reformats its solution into JSON (Phase~2), with an explicit separator. This decouples reasoning from format compliance, allowing us to test the capacity competition hypothesis.

\subsection{Evaluation}

We evaluate by exact match on the extracted numerical answer. For JSON responses, we first attempt to parse the JSON and extract the \texttt{final\_answer} field; for all responses, we fall back to extracting the last number via regex. We use temperature~0 (deterministic) for all primary experiments; McNemar tests and reported accuracies are computed on this deterministic run. To assess replicability, we additionally ran two stochastic reruns (temperature 0.3) for key experiments (MATH-Hard, delayed structure, schema gradient). Reported confidence intervals reflect variability over one deterministic run plus two stochastic reruns; per-temperature estimates differed by only 1--3pp, well within the reported intervals and far smaller than the observed effect sizes (28--36pp). All experiments use the same problems across conditions to enable paired McNemar comparisons.

\paragraph{Batch structure and cross-table consistency.} The delayed-structure ablation (Table~\ref{tab:delayed}), schema gradient (Table~\ref{tab:gradient}), and some cross-benchmark cells (Table~\ref{tab:crossbench}) were run as independent paired experiments at different times. The CoT rows in those tables are table-local controls and should not be compared as absolute estimates across tables; minor variation in absolute baselines (up to $\sim$6pp for GPT-4o-mini) reflects API nondeterminism across runs. Within each table, all conditions were evaluated on the same item set, scoring pipeline, and API snapshot, ensuring valid paired inference. Our reported deltas, recovery percentages, and McNemar tests are computed within-batch. The canonical CoT baseline is established in Table~\ref{tab:main} (3-run CI).

\section{Results}
\label{sec:results}

\subsection{The Capacity-Dependent Effect}
\label{sec:capacity-dependent}

\begin{figure}[t]
\centering
\includegraphics[width=\columnwidth]{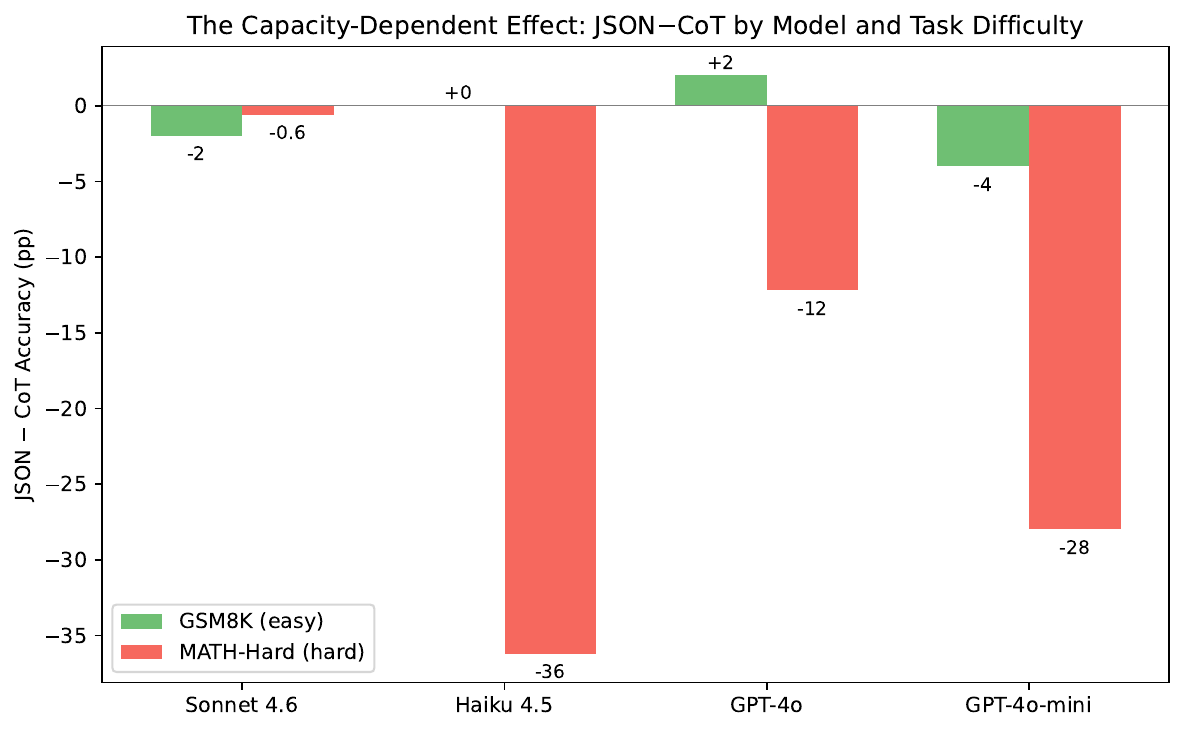}
\caption{JSON--CoT accuracy difference (\%) across 4 models on GSM8K (easy, single run) vs.\ MATH-Hard (hard, 3-run means). On easy tasks, JSON effects are small ($\pm$4\%). On hard tasks, the effect diverges: neutral for Sonnet ($-$0.6pp), severely negative for Haiku ($-$36pp) and GPT-4o-mini ($-$28pp).}
\label{fig:hero}
\end{figure}

Table~\ref{tab:main} presents accuracy across all models, benchmarks, and conditions, and Figure~\ref{fig:hero} visualizes the JSON--CoT accuracy difference. The central finding is that structured output effects reverse direction between easy and hard tasks.

\begin{table}[t]
\centering
\small
\begin{tabular}{@{}llccc@{}}
\toprule
& & \multicolumn{3}{c}{\textbf{Output Condition}} \\
\cmidrule(l){3-5}
\textbf{Bench.} & \textbf{Model} & CoT & JSON & XML \\
\midrule
\multirow{4}{*}{\rotatebox{90}{\small GSM8K}}
& Sonnet 4.6   & 97.0 & 95.0 & 94.0 \\
& Haiku 4.5    & 96.0 & 96.0 & 96.0 \\
& GPT-4o       & 90.0 & 92.0 & 93.0 \\
& GPT-4o-mini  & 95.0 & 91.0 & 87.0 \\
\midrule
\multirow{4}{*}{\rotatebox{90}{\small MATH-H}}
& Sonnet 4.6   & 89.3\tinypm{1.7} & 88.7\tinypm{4.0} & 89.3\tinypm{0.7} \\
& Haiku 4.5    & 88.7\tinypm{0.7} & \underline{52.5\tinypm{2.5}} & \underline{58.7\tinypm{1.3}} \\
& GPT-4o       & 56.2\tinypm{1.5} & \underline{44.0\tinypm{4.4}} & \underline{43.0} \\
& GPT-4o-mini  & 62.3\tinypm{4.0} & \underline{34.3\tinypm{1.7}} & \underline{34.3\tinypm{3.3}} \\
\bottomrule
\end{tabular}
\caption{Accuracy (\%) on GSM8K (easy, single run) and MATH-Hard Level~5 (hard, mean $\pm$ 95\% CI over 3 runs; GPT-4o XML is single run). \underline{Underline}: $>$5pp below CoT. On GSM8K, JSON effects are within $\pm$4pp (max format effect across all conditions is 8pp for GPT-4o-mini XML). On MATH-Hard, JSON is neutral for Sonnet but severely harms Haiku ($-$36.2pp) and GPT-4o-mini ($-$28.0pp). Detailed prose control results are in Table~\ref{tab:decompose}.}
\label{tab:main}
\end{table}

\paragraph{Easy tasks (GSM8K).} All models achieve 87--97\% across all conditions. The maximum effect of any format is 8.0 percentage points (GPT-4o-mini CoT 95.0\% vs.\ XML 87.0\%). No format produces a statistically significant change for any model ($p > 0.05$ for all pairwise comparisons). Structured output is essentially neutral when tasks are within a model's comfort zone.

\paragraph{Hard tasks (MATH-Hard).} The picture changes dramatically. For Sonnet~4.6, JSON accuracy (88.7\%) is statistically indistinguishable from CoT (89.3\%, $p = 0.25$). For weaker models, however, JSON causes severe degradation: Haiku drops from 88.7\% to 52.5\% ($-$36.2pp, $p < 0.0001$) and GPT-4o-mini drops from 62.3\% to 34.3\% ($-$28.0pp, $p = 0.0002$).

This constitutes a \emph{capacity-dependent divergence}: the same structured format that is neutral on easy tasks becomes severely harmful on hard tasks, but only for models operating near their capability limits.

\subsection{Format-Specific vs.\ Prompt Length Effects}
\label{sec:format-specific}

A natural concern is that the degradation arises from longer, more complex prompts rather than the structured format itself. Our detailed prose (DP) control addresses this by providing equally detailed analytical instructions without any format constraint.

\begin{figure}[t]
\centering
\includegraphics[width=\columnwidth]{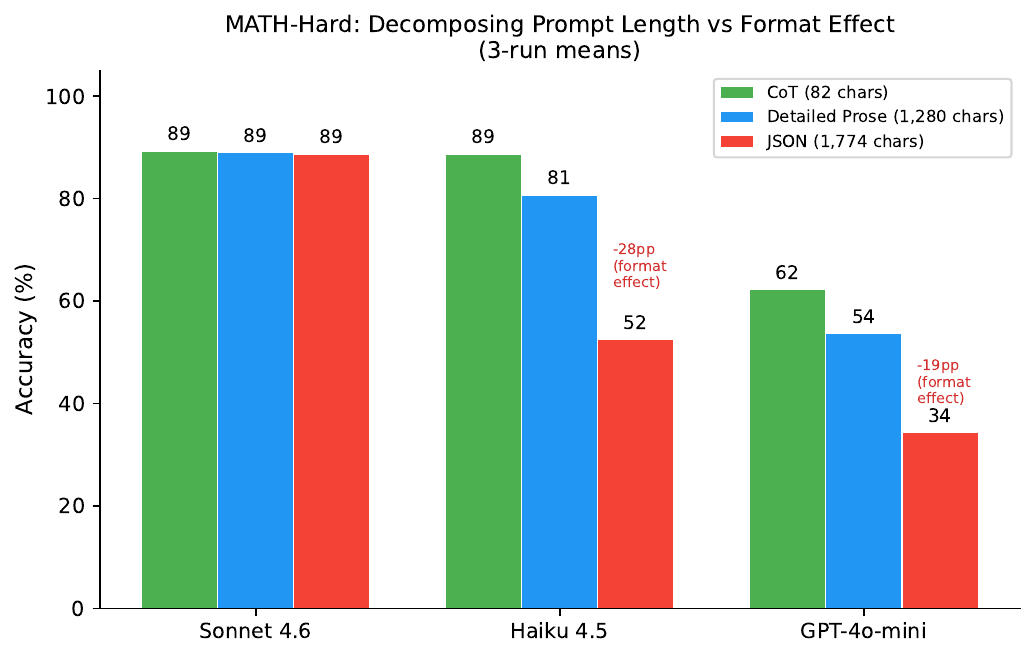}
\caption{Decomposing prompt length vs.\ format effect on MATH-Hard (3-run means). The gap between Detailed Prose (blue) and CoT (green) measures prompt-length cost; the gap between JSON (red) and Detailed Prose measures format-specific cost. For Haiku, the format-specific gap is 28 percentage points. GPT-4o is omitted (DP CI data unavailable).}
\label{fig:decompose}
\end{figure}

Table~\ref{tab:decompose} decomposes the total effect into prompt-length and format-specific components, and Figure~\ref{fig:decompose} visualizes the three conditions.

\begin{table}[h!]
\centering
\small
\begin{tabular}{@{}lcccc@{}}
\toprule
\textbf{Model} & CoT & DP & JSON & $\Delta_{\text{struct}}$ \\
\midrule
Sonnet 4.6  & 89.3 & 89.0 & 88.7 & $-$0.3 \\
Haiku 4.5   & 88.7 & 80.7 & 52.5 & $-$28.2*** \\
GPT-4o      & 56.2 & --- & 44.0 & --- \\
GPT-4o-mini & 62.3 & 53.7 & 34.3 & $-$19.3** \\
\bottomrule
\end{tabular}
\caption{Decomposing the structured output effect on MATH-Hard (3-run means). $\Delta_{\text{struct}}$ = JSON $-$ DP isolates the format-specific component after controlling for prompt length. GPT-4o DP is omitted (CI data unavailable). *** $p < 0.001$, ** $p < 0.01$.}
\label{tab:decompose}
\end{table}

The prompt-length effect (DP vs.\ CoT) is modest: $-$0.3pp for Sonnet and $-$8.0pp for Haiku, suggesting that detailed instructions alone account for a small part of the observed degradation. The format-specific effect (JSON vs.\ DP) is where models diverge: $-$0.3pp for Sonnet (neutral) but $-$28.2pp for Haiku ($p < 0.0001$) and $-$19.3pp for GPT-4o-mini ($p < 0.01$).

This makes prompt length alone an unlikely explanation for the observed degradation. The harm is attributable to premature serialization---simultaneously reasoning and producing schema-compliant output---not to the informational content of the instructions.

\subsection{Cross-Format Consistency}
\label{sec:crossformat}

To verify that the effect is not specific to JSON parsing, we compare JSON and XML on MATH-Hard. Both formats show consistent degradation for weaker models: Haiku drops $-$36.2pp with JSON and $-$30.0pp with XML; GPT-4o drops $-$12.2pp with JSON and $-$13.2pp with XML; GPT-4o-mini drops $-$28.0pp with JSON and $-$28.0pp with XML (all $p < 0.01$ vs.\ CoT). The consistency across two syntactically different formats suggests the effect arises from the general burden of format compliance, not from idiosyncratic properties of any single format.

\subsection{Production Structured Output Modes}
\label{sec:jsonmode}

To test whether the observed tax extends to production-style structured output, we compare four structured-output mechanisms on GPT-4o-mini MATH-Hard (Table~\ref{tab:prodmodes}).

\begin{table}[t]
\centering
\small
\begin{tabular}{@{}lccc@{}}
\toprule
\textbf{Condition} & \textbf{Acc.} & \textbf{Tokens} & \textbf{vs.\ CoT} \\
\midrule
CoT (free-form) & 60\% & 752 & --- \\
API JSON mode & 51\% & 436 & $-$9 \\
Instruction JSON (heavy) & 34\% & 852 & $-$26*** \\
Function calling (forced) & 10\% & 328 & $-$50*** \\
\bottomrule
\end{tabular}
\caption{GPT-4o-mini MATH-Hard across production structured-output modes. API JSON mode (\texttt{response\_format=json\_object}) shows a nonsignificant $-$9pp gap vs.\ CoT ($p = 0.12$). Instruction-prompted heavy JSON is significantly worse than API JSON mode ($p = 0.0095$). Forced function calling (\texttt{tool\_choice=forced}) collapses accuracy to 10\%: under forced submission, models in practice emit only tool-call arguments without reasoning content, leaving no visible deliberation surface. *** $p < 0.001$ vs.\ CoT.}
\label{tab:prodmodes}
\end{table}

The results reveal a spectrum of reasoning freedom. API JSON mode imposes a light constraint (valid JSON required, no schema) and shows a modest, nonsignificant accuracy drop with \emph{fewer} tokens than CoT (436 vs.\ 752). Instruction-prompted JSON with a 1,774-character schema is significantly worse than API JSON mode (34\% vs.\ 51\%; McNemar $p = 0.0095$), attributable to redundant field requirements rather than JSON syntax. GPT-4o shows a consistent pattern: CoT 56\%, API JSON mode 51\%, instruction JSON 40\%.

Forced function calling is the most constrained mode: under \texttt{tool\_choice=forced}, models in practice emit only tool-call arguments without preceding reasoning content (median 328 tokens vs.\ 752 for CoT), so reasoning must be compressed into structured parameters. Accuracy drops to 10\%---a production-interface bottleneck rather than a format effect per se. Under \texttt{tool\_choice=auto}, GPT-4o-mini avoids the tool entirely (only 8/100 used it, achieving 57\%$\approx$CoT), while GPT-4o learned to emit both content and tool calls simultaneously (70/100 responses included both, achieving 57\%$\approx$CoT 56\%), suggesting that coordinating reasoning and tool submission is itself a capacity-dependent skill.

\subsection{Delayed Structure Ablation}
\label{sec:delayed-results}

The results above suggest that weaker models struggle to simultaneously reason \emph{and} comply with format constraints. We test this capacity competition hypothesis with the delayed structure condition (Section~\ref{sec:delayed}), where models reason freely first, then reformat into JSON.

\begin{table}[t]
\centering
\small
\begin{tabular}{@{}lcccc@{}}
\toprule
\textbf{Model} & CoT & Delayed & Simult. & Recovery \\
\midrule
Sonnet 4.6  & 91.0 & 94.0 & 92.0 & --- \\
Haiku 4.5   & 89.0\tinypm{2.0} & 82.0\tinypm{2.3} & 54.3\tinypm{5.8} & 79.5\% \\
GPT-4o      & 53.0\tinypm{2.3} & 58.3\tinypm{1.3} & 42.0 & $>$100\% \\
GPT-4o-mini & 59.6\tinypm{6.3} & 56.7\tinypm{2.4} & 37.4\tinypm{4.0} & 86.9\% \\
\bottomrule
\end{tabular}
\caption{Delayed structure ablation on MATH-Hard (3-run means $\pm$ 95\% CI where available; Sonnet and GPT-4o Simult.\ are single-run). ``Recovery'' = (Delayed $-$ Simult.) / (CoT $-$ Simult.) $\times$ 100\%. Delayed formatting recovers 80\% (Haiku) and 87\% (GPT-4o-mini) of lost performance. GPT-4o consistently exceeds CoT under delayed structure, likely reflecting scaffolding benefit from the two-phase prompt.}
\label{tab:delayed}
\end{table}

Table~\ref{tab:delayed} shows that delayed formatting recovers most of the performance lost to simultaneous formatting. For Haiku, accuracy rises from 54.3\% (simultaneous) to 82.0\% (delayed), recovering 79.5\% of the gap to CoT (89.0\%) under 3-run replication. For GPT-4o, the delayed prompt consistently exceeds CoT across three runs (mean 58.3\% vs.\ 53.0\%), yielding apparent recovery $>$100\%. Since the delayed prompt (522 chars, explicit ``Phase 1: THINK / Phase 2: FORMAT'') is more structured than the minimal CoT prompt (82 chars), this gain may reflect beneficial task scaffolding rather than delayed serialization alone. Because delayed structure necessarily changes the instruction sequence, we do not rely on this condition alone; the same-prompt token-budget ablation (Section~\ref{sec:analysis}) provides an independent test of capacity pressure without any prompt change.

\subsection{Generalization Beyond Math}
\label{sec:generalization}

\begin{table}[t]
\centering
\small
\begin{tabular}{@{}lccccc@{}}
\toprule
\textbf{Model} & \rotatebox{70}{GSM8K} & \rotatebox{70}{MATH-H} & \rotatebox{70}{BBH-Log} & \rotatebox{70}{BBH-Trk} & \rotatebox{70}{MMLU-P} \\
\midrule
Sonnet 4.6  & $-$2.0 & +5.0  & +2.0   & +1.0   & +1.3 \\
Haiku 4.5   & +0.0  & \underline{$-$34.0} & $-$3.0  & $-$1.0  & +2.3 \\
GPT-4o      & +2.0  & \underline{$-$16.0}  & $-$5.0  & $-$1.0  & \underline{$-$10.7} \\
GPT-4o-mini & $-$4.0 & \underline{$-$26.0} & $-$8.0 & \underline{$-$24.0} & \underline{$-$19.3} \\
\bottomrule
\end{tabular}
\caption{JSON--CoT accuracy difference (pp) across 5 benchmarks (single run; MATH-H values differ slightly from the 3-run means in Table~\ref{tab:main}). \underline{Underline}: $>$10pp degradation. GPT-4o-mini is consistently harmed across all non-easy benchmarks and domains. Sonnet is consistently unaffected under 3-run replication (Table~\ref{tab:main}); the +5.0 here is single-run noise. Haiku shows task-dependent effects: severely harmed on math but neutral or positive on other domains.}
\label{tab:crossbench}
\end{table}

To test whether these findings generalize beyond math, we evaluate on two additional benchmark families: BIG-Bench Hard (logical deduction and object tracking) and MMLU-Pro (law, physics, philosophy). Table~\ref{tab:crossbench} shows the JSON--CoT accuracy difference across all five benchmarks.

The pattern from MATH-Hard generalizes: GPT-4o-mini suffers large degradation across all domains ($-$8 to $-$26pp). Sonnet remains neutral or positive on every benchmark (the +5.0 on MATH-Hard is single-run noise; 3-run replication shows $-$0.6pp, Table~\ref{tab:main}). Two nuances emerge. First, Haiku is severely harmed on MATH-Hard ($-$34pp) but slightly \emph{helped} on MMLU-Pro (+2.3pp)---a per-category analysis reveals that JSON helps Haiku on law (+15pp, where option-by-option analysis acts as scaffolding) but hurts on physics ($-$13pp, which requires sequential derivation). Second, the effect tracks task difficulty relative to each model: benchmarks where a model scores $>$90\% under CoT show negligible format effects, while those below 70\% show the largest degradation.

\paragraph{Formal interaction test.} To test the format$\times$capability interaction, we fit a unified item-level logistic regression across all models and hard benchmarks. We excluded model-item pairs where either condition produced a non-evaluable response due to API content-filter termination, server-side rejection, or absence of an extractable final answer despite a completed reasoning trace (14 pairs, 28 rows; both format observations removed per pair to preserve paired comparability). The core model includes the 4 main models $\times$ 4 benchmarks ($N = 4{,}772$), with frontier probes (Sonnet and Opus on AIME, Opus on GPQA) as exploratory out-of-sample extensions:

\begin{table}[t]
\centering
\small
\begin{tabular}{@{}lcc@{}}
\toprule
\textbf{Analysis} & \textbf{Statistic} & \textbf{$p$-value} \\
\midrule
Item-level logistic (std.\ SE) & $z = 4.62$ & $3.9 \times 10^{-6}$ \\
Clustered by cell ($n = 16$) & $z = 2.90$ & $3.8 \times 10^{-3}$ \\
Cell-level Spearman ($n = 16$) & $r_s = 0.50$ & $0.047$ \\
+Frontier cells ($n = 19$) & $r_s = 0.51$ & $0.026$ \\
\midrule
Efficiency proxy (item-level) & $z = -4.01$ & $6.2 \times 10^{-5}$ \\
\bottomrule
\end{tabular}
\caption{Interaction between format (JSON vs.\ CoT) and model capacity. The top rows use cross-validated CoT accuracy as the capacity proxy. Adding 3 frontier cells (Sonnet AIME, Opus AIME, Opus GPQA) maintains significance. The bottom row uses reasoning efficiency (log median CoT tokens) as an alternative proxy.}
\label{tab:interaction}
\end{table}

\begin{figure}[t]
\centering
\includegraphics[width=\columnwidth]{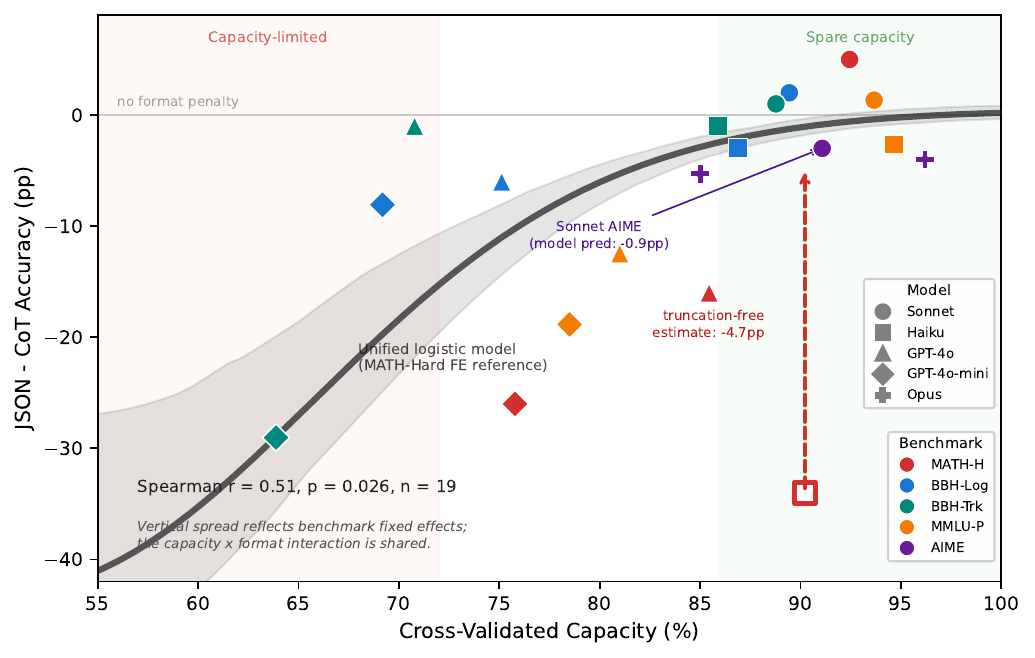}
\caption{Format penalty (JSON$-$CoT accuracy) vs.\ cross-validated capacity across 19 model$\times$benchmark cells. Shaded regions indicate capacity regimes: models in the comfort zone ($>$86\%) show negligible format effects, while capacity-limited models ($<$72\%) suffer large penalties. Haiku on MATH-Hard is an outlier driven by token-budget truncation. Spearman $r_s = 0.51$, $p = 0.026$.}
\label{fig:capacity}
\end{figure}

\vspace{-0.5em}
\begin{equation*}
\text{correct} \sim \text{is\_json} \times \text{capacity} + \text{benchmark}
\end{equation*}
\vspace{-0.5em}

\noindent where \emph{capacity} is cross-validated: each model's mean CoT accuracy on the \emph{other} hard benchmarks, avoiding circularity. The interaction is significant (Table~\ref{tab:interaction}), holding under clustered standard errors ($p = 3.8 \times 10^{-3}$) and directionally consistent at the cell level ($n = 16$, Spearman $r_s = 0.50$, $p = 0.047$). Adding three frontier cells---Sonnet on AIME ($-$3.0pp), Opus on AIME ($-$5.3pp), and Opus on GPQA ($-$4.0pp)---maintains significance ($n = 19$, $r_s = 0.51$, $p = 0.026$). As a sensitivity check, replacing Haiku's truncation-inflated MATH-Hard tax ($-$34.0pp) with its truncation-free estimate ($-$4.7pp) strengthens the relationship ($r_s = 0.67$, $p = 0.004$). Figure~\ref{fig:capacity} visualizes this relationship across all 19 cells.

An alternative model using reasoning efficiency---log median CoT tokens on matched correct items---as the capacity proxy supports the same direction ($p = 6.2 \times 10^{-5}$): models that need more tokens to solve the same problems under CoT suffer larger JSON penalties. On the easy benchmark (GSM8K), the interaction is not significant ($p = 0.34$), consistent with our thesis that format effects emerge only near capacity boundaries.

\section{Analysis and Discussion}
\label{sec:analysis}

\subsection{Why Structure Does Not Hurt Strong Models}

Sonnet~4.6 shows no degradation under JSON on MATH-Hard (88.7\% vs.\ 89.3\% CoT, $p = 0.25$). Under 3-run replication, Sonnet scores $\sim$89\% across all conditions---CoT, JSON, XML, and detailed prose---with no significant differences. We interpret this as strong models having sufficient spare capacity to absorb the format-compliance burden without displacing reasoning. Note that small positive JSON effects observed in single runs (+1--3pp on BBH and MMLU-Pro) do not replicate reliably and should not be interpreted as JSON enhancing reasoning.

\subsection{Why Structure Hurts Weak Models}

For Haiku~4.5 and GPT-4o-mini, simultaneous formatting causes dramatic degradation (up to $-$36pp). Our delayed-structure ablation (Section~\ref{sec:delayed-results}) shows that deferring formatting to after reasoning recovers most of the lost accuracy (3-run mean: 80\% for Haiku and 87\% for GPT-4o-mini). GPT-4o consistently exceeds its CoT baseline under delayed prompting (3-run mean 58.3\% vs.\ 53.0\%), suggesting the two-phase prompt provides scaffolding beyond mere serialization delay. The models possess the mathematical knowledge---it is not absent but \emph{inaccessible} when generation capacity is diverted to format compliance.

We hypothesize \textbf{capacity competition}: producing valid JSON requires tracking bracket nesting, field ordering, type constraints, and quote escaping---operations that share the same autoregressive generation process as mathematical reasoning. For capacity-limited models, these format-compliance operations crowd out reasoning steps.

\paragraph{Schema complexity gradient.} To test this hypothesis, we vary schema complexity across four levels while keeping the task fixed (MATH-Hard, 100 problems): \emph{minimal} (answer-only JSON, 73 chars), \emph{light} (one reasoning field, 117 chars), \emph{medium} (structured steps, 177 chars), and \emph{heavy} (our full nested schema, 1,774 chars). If capacity competition is the mechanism, degradation should increase monotonically with schema burden for weaker models.

\begin{table}[t]
\centering
\small
\begin{tabular}{@{}lccccc@{}}
\toprule
\textbf{Model} & CoT & Min. & Light & Med. & Heavy \\
\midrule
Sonnet 4.6  & 92.3\tinypm{2.4} & 91.7\tinypm{9.5} & 95.7\tinypm{4.6} & 93.7\tinypm{4.0} & 93.0$^\dagger$ \\
Haiku 4.5   & 86.0\tinypm{3.0} & 89.7\tinypm{3.6} & 90.3\tinypm{0.7} & 86.0\tinypm{1.1} & \underline{55.3\tinypm{3.6}} \\
GPT-4o      & 56.3\tinypm{1.7} & \underline{17.0\tinypm{2.3}} & 58.7\tinypm{1.7} & 54.3\tinypm{1.3} & \underline{42.3\tinypm{3.5}} \\
GPT-4o-mini & 59.0\tinypm{3.9} & 53.7\tinypm{4.0} & 58.3\tinypm{2.6} & \underline{47.1\tinypm{5.8}} & \underline{36.3\tinypm{3.5}} \\
\bottomrule
\end{tabular}
\caption{Schema complexity gradient on MATH-Hard (3-run means $\pm$ 95\% CI except Sonnet heavy, which retains the original within-batch value). Among schemas that elicit reasoning (light--heavy), weaker models show monotonic degradation with increasing schema complexity. Minimal (answer-only) is a special case: it suppresses reasoning rather than competing with it, explaining GPT-4o's collapse to 17\%. $^\dagger$A later replication audit of Sonnet heavy produced 60\% and 63\%, while all other Sonnet conditions remained stable; we attribute this to API serving-layer drift rather than a format effect.}
\label{tab:gradient}
\end{table}

Table~\ref{tab:gradient} shows a clear dose-response pattern among schemas that include reasoning fields (light through heavy). For Haiku, 3-run mean accuracy drops from 90.3\% (light) to 86.0\% (medium) to 55.3\% (heavy). GPT-4o-mini shows a similar gradient: 58.3\%$\to$47.1\%$\to$36.3\% (light$\to$heavy). In the original experimental batch, Sonnet remained stable across all levels; however, a later replication audit of the heavy condition produced substantially lower scores (60\%, 63\%) while cot/light/medium remained stable, suggesting API serving-layer drift rather than a genuine format effect.

The \emph{minimal} schema (answer-only JSON) is a distinct phenomenon: it suppresses chain-of-thought reasoning rather than competing with it. GPT-4o's collapse to 17\% under minimal---while recovering to 59\% under light---shows that removing the reasoning field is harmful for a different reason. We therefore restrict the capacity competition claim to reasoning-bearing schemas.

Importantly, \textbf{no parse failures occurred} among successfully generated responses (0\% extraction failure rate). A small number of items (14 model-item pairs, $<$0.6\%) were excluded from the interaction analysis because either condition produced a non-evaluable response due to API content filtering or server-side rejection unrelated to model reasoning. All accuracy differences reflect genuine reasoning errors, not JSON formatting or answer extraction artifacts.

\paragraph{Failure analysis.} We classified all 81 items where a model answered correctly under CoT but incorrectly under heavy JSON on MATH-Hard (35 Haiku, 25 GPT-4o-mini, 21 GPT-4o). Table~\ref{tab:errors} shows the breakdown.

\begin{table}[t]
\centering
\small
\begin{tabular}{@{}lcccc@{}}
\toprule
\textbf{Error Type} & Haiku & 4o-mini & GPT-4o & \textbf{Total} \\
\midrule
Reasoning omission & 20 & 8 & 2 & 30 (37\%) \\
Arithmetic error   & 3 & 7 & 12 & 22 (27\%) \\
Wrong approach     & 7 & 8 & 6 & 21 (26\%) \\
Premature answer   & 4 & 1 & 1 & 6 (7\%) \\
Nonsense           & 1 & 1 & 0 & 2 (2\%) \\
\bottomrule
\end{tabular}
\caption{Error classification for 81 items correct under CoT but wrong under heavy JSON on MATH-Hard. The error profile shifts with capability: Haiku is dominated by reasoning omissions (57\%), while GPT-4o skews toward arithmetic errors (57\%).}
\label{tab:errors}
\end{table}

The model-level pattern is revealing: Haiku's failures are overwhelmingly reasoning omissions (57\%), suggesting it stops the derivation prematurely to fill JSON fields. GPT-4o retains the correct approach but makes more arithmetic mistakes (57\% of its errors), consistent with reduced precision under divided attention. This supports capacity competition: heavy JSON does not corrupt the output format---it diverts effort from completing the mathematical derivation.

Notably, Haiku and Sonnet have near-identical CoT baselines on MATH-Hard (85--90\%), yet Haiku collapses under heavy JSON while Sonnet does not. This suggests models differ not only in reasoning ability but in \emph{surplus capacity}---the headroom available for additional generation constraints beyond the core task.

\subsection{Token-Budget Evidence for Capacity Competition}

If capacity competition is real, it should be visible in the generation budget. We examine three lines of token-level evidence.

\paragraph{Reasoning efficiency.} On MATH-Hard, Sonnet and Haiku achieve comparable CoT accuracy (87\%), but Sonnet solves problems in far fewer tokens. Comparing only the 77 items both models answer correctly, Sonnet uses a median of 94 tokens while Haiku uses 665 (7.1$\times$; Wilcoxon $p = 2.5 \times 10^{-14}$). This pattern holds across all benchmarks: BBH-Log 116 vs.\ 527 tokens (4.5$\times$, $p = 3.8 \times 10^{-17}$, $n = 94$), BBH-Trk 171 vs.\ 430 (2.5$\times$, $p = 8.3 \times 10^{-18}$, $n = 98$), MMLU-Pro 4 vs.\ 392 (97.9$\times$, $p = 2.0 \times 10^{-32}$, $n = 194$). We interpret this as an independent proxy for \emph{effective capacity margin}: Sonnet reasons compactly, leaving ample headroom for schema overhead; Haiku consumes most of its generation budget on reasoning alone.

\paragraph{Truncation as a capacity bottleneck.} Under heavy JSON with \texttt{max\_tokens}$=$2048, Haiku hits the token limit on 49\% of responses (45 of 49 truncated responses are incorrect). By contrast, Sonnet hits the limit on only 5\%, and GPT-4o-mini on 0\%. Yet GPT-4o-mini still loses $-$28pp under JSON---demonstrating that truncation is \emph{one} mechanism of capacity competition, but not the only one.

\paragraph{Token-budget ablation.} To isolate the truncation contribution, we reran Haiku's heavy JSON on MATH-Hard with increased token budgets (Table~\ref{tab:budget}).

\begin{table}[t]
\centering
\small
\begin{tabular}{@{}lccc@{}}
\toprule
\textbf{Condition} & \textbf{Accuracy} & \textbf{Truncation} & \textbf{vs.\ CoT} \\
\midrule
CoT (2048) & 88.7\tinypm{0.7} & 0\% & --- \\
JSON (2048) & 52.5\tinypm{2.5} & 49\% & $-$36.2 \\
JSON (4096) & 81.7\tinypm{2.8} & 9\% & $-$7.0 \\
JSON (8192) & 84.0\tinypm{3.4} & 2\% & $-$4.7 \\
\bottomrule
\end{tabular}
\caption{Haiku token-budget ablation on MATH-Hard (\%). Increasing \texttt{max\_tokens} from 2048 to 8192 recovers most of the JSON penalty (52.5\%$\to$84.0\%), but a residual gap of $\sim$4.7pp remains relative to CoT (88.7\%). The 8192 result reports mean $\pm$ 95\% CI over 3 runs (temperature 0.3).}
\label{tab:budget}
\end{table}

Increasing the budget from 2048 to 8192 recovers 87\% of the original gap (52.5\%$\to$84.0\%), indicating that truncation is the dominant proximate mechanism for Haiku. However, two observations suggest the phenomenon is not reducible to truncation alone: (i) a residual $\sim$4.7pp gap persists at 8192; and (ii) GPT-4o-mini shows no recovery under the same intervention (34\%$\to$36\% at 8192, 0\% truncation at both budgets), consistent with a distinct non-truncation capacity cost.

These results support a layered account: structured formatting taxes capacity through \emph{multiple channels}---output-budget exhaustion for models with low reasoning efficiency, and reasoning-allocation interference even when the output fits within the budget.

\paragraph{Frontier model exploration.} To probe whether the effect attenuates or disappears at frontier scale, we evaluated Opus~4.7 on two hard benchmarks: AIME 2020--2024 (133 competition math problems) and GPQA Diamond (198 PhD-level science questions). Table~\ref{tab:frontier} summarizes the results.

\begin{table}[t]
\centering
\small
\begin{tabular}{@{}lrrccc@{}}
\toprule
\textbf{Benchmark} & $N$ & \textbf{Tax} & \textbf{CoT$>$J} & \textbf{J$>$CoT} & $p$ \\
\midrule
AIME & 133 & $-$5.3pp & 10 & 3 & 0.092 \\
GPQA Diamond & 198 & $-$4.0pp & 17 & 9 & 0.169 \\
\midrule
Pooled (expl.) & 331 & $-$4.5pp & 27 & 12 & 0.024 \\
\midrule
HLE (floor) & 500 & $-$0.2pp & 41 & 40 & 1.0 \\
\bottomrule
\end{tabular}
\caption{Frontier format tax: Opus~4.7 on hard benchmarks. Individual benchmarks show consistent directional penalties but are individually underpowered. An exploratory pooled analysis across AIME and GPQA reaches significance ($p = 0.024$). HLE serves as a floor-effect control: at 28.8\% CoT accuracy, format effects collapse into symmetric noise.}
\label{tab:frontier}
\end{table}

Neither AIME nor GPQA individually reaches significance, but both show a consistent $-$4 to $-$5pp penalty with a 2:1 ratio of CoT-only to JSON-only discordant pairs. An exploratory pooled analysis yields $p = 0.024$ (labeled post-hoc); including HLE dilutes the effect ($p = 0.17$), consistent with the capacity-regime account: format penalties are detectable near the capability boundary but vanish at the floor.

\paragraph{Extended thinking.} As an exploratory probe, we enabled Opus~4.7's extended thinking mode on all 133 AIME problems. JSON accuracy rose from 91.0\% (standard) to 94.7\% (extended thinking), narrowing the format tax from $-$5.3pp to $-$2.3pp relative to CoT (McNemar $p = 0.27$, not significant). This is suggestive evidence that model-internal reasoning phases may reduce capacity competition.

\subsection{Reasoning-Formatting Separation}

Across all our experiments, a consistent principle emerges: structured output is harmful primarily when it is imposed \emph{during} reasoning. We formalize this as follows.

\noindent\textbf{Premature serialization} occurs when an interface requires a model to produce schema-compliant tokens before or during its primary reasoning process. This couples two objectives that can compete under limited capacity: solving the task and maintaining a valid external representation. Its counterpart, \textbf{delayed serialization}, separates deliberation from formatting---the model reasons freely first, then serializes the result.

We define \emph{reasoning freedom} as the degree to which an interface allows a model to allocate tokens and attention to task reasoning before satisfying serialization constraints. Table~\ref{tab:spectrum} orders all tested mechanisms by reasoning freedom.

\begin{table}[t]
\centering
\small
\begin{tabular}{@{}lcccc@{}}
\toprule
\textbf{Mechanism} & \textbf{Free reasoning?} & \textbf{Haiku} & \textbf{4o} & \textbf{4o-mini} \\
\midrule
CoT (free-form) & Full & 87 & 56 & 60 \\
Claude FC (text+tool) & Full (native) & 88 & --- & --- \\
Two-round tool & Yes (multi-turn) & 86 & 51 & 59 \\
GPT-4o auto FC & Yes (learned) & --- & 57 & --- \\
Delayed structure & Yes (prompted) & 84 & 59 & 55 \\
\midrule
API JSON mode & Partial & --- & 51 & 51 \\
Instruction JSON & No (simultaneous) & 53 & 40 & 34 \\
Forced FC & None & --- & --- & 10 \\
\bottomrule
\end{tabular}
\caption{Reasoning freedom spectrum on MATH-Hard (\%). Mechanisms above the midline preserve a reasoning phase before structured submission; those below do not. Performance tracks reasoning freedom consistently across models. FC = function calling.}
\label{tab:spectrum}
\end{table}

The pattern is consistent: mechanisms that preserve reasoning freedom (CoT, Claude text+tool, two-round orchestration, delayed structure) yield accuracy near baseline, while mechanisms that impose premature serialization (heavy instruction JSON, forced function calling) produce large drops. GPT-4o learned to emit both content and tool calls simultaneously under \texttt{tool\_choice=auto} (70/100 responses included both), while GPT-4o-mini did not (8/100)---suggesting that this coordination is itself a capacity-dependent skill.

\vspace{0.5em}
\noindent\fbox{\parbox{0.95\linewidth}{
\textbf{Reasoning-Formatting Separation Principle.} For capacity-limited reasoning tasks, structured-output interfaces should preserve an unconstrained deliberation phase before requiring schema-compliant output. The expected format tax increases as reasoning freedom decreases and as task demand approaches model capacity.
}}
\vspace{0.5em}

\paragraph{Design implications.} This principle translates into concrete API and workflow recommendations:
\begin{enumerate}
    \item \textbf{Prefer delayed serialization} over simultaneous reasoning and formatting when task difficulty is uncertain.
    \item \textbf{Allow text alongside tool calls} (as Claude's API does natively) rather than forcing content-free tool-call-only responses.
    \item \textbf{Keep schemas lightweight} when simultaneous formatting is unavoidable; minimize redundant fields that do not contribute to reasoning.
    \item \textbf{Treat forced function calling as risky} for hard reasoning tasks unless the model has a separate reasoning channel (e.g., extended thinking).
\end{enumerate}

\noindent As a practical decision rule:
\begin{itemize}
    \item \textbf{Easy tasks} (model accuracy $>$90\% without format): structured output is safe.
    \item \textbf{Hard tasks, strong models}: structured output is neutral.
    \item \textbf{Hard tasks, weaker models}: use delayed formatting---let the model reason freely, then reformat in a second pass. This preserves both structured output and reasoning quality.
\end{itemize}

\subsection{Limitations}

\paragraph{Schema complexity confounds.} Our gradient varies prompt length, field count, and nesting depth simultaneously; future work should isolate each factor.

\paragraph{Math-heavy evaluation.} The strongest effects appear on math tasks. Capacity competition may manifest differently on generation or open-ended tasks.

\paragraph{Single-run cells.} The cross-benchmark table (BBH, MMLU-Pro, GSM8K columns) and some frontier experiments use single runs. Large effect sizes ($>$20pp) and paired McNemar tests provide adequate power, but confidence intervals are available only for MATH-Hard, the delayed-structure ablation, and the schema gradient.

\paragraph{Frontier CV instability.} Opus~4.7 was evaluated on only two hard benchmarks (AIME, GPQA), so its cross-validated capacity is estimated from a single held-out benchmark rather than an average over three. The core 16-cell analysis (4 main models $\times$ 4 benchmarks) is not affected; Opus cells are exploratory extensions.

\paragraph{No grammar-constrained decoding.} We test instruction-prompted schemas, API-native JSON mode, and tool-calling interfaces, but not grammar-constrained decoding or strict schema decoders that enforce validity at the token level.

\paragraph{Closed-source models only.} Mechanistic investigation would require open-weight models and access to internal representations. Closed-API models are also subject to serving-layer changes: a replication audit of the Sonnet heavy-schema condition produced 60--63\% accuracy (vs.\ 93\% in the original batch), while all other Sonnet conditions remained stable, suggesting backend drift rather than a format effect. We retain within-batch values for all tables and rely on within-batch paired comparisons.

\paragraph{Frontier boundary gap.} No public math benchmark places Opus in the 50--70\% capacity boundary zone where our theory predicts the largest effects. On HLE (500 math problems), CoT accuracy is 28.8\% with symmetric discordance (McNemar $p = 1.0$), consistent with a floor regime. Constructing a calibrated boundary subset is an important direction for future work.

\section{Conclusion}
\label{sec:conclusion}

Our results suggest that structured output format effects on LLM reasoning are not an inherent penalty of machine-readable formats, but arise from \emph{premature serialization}---forcing models to produce schema-compliant tokens before they have completed reasoning. Across instruction-prompted JSON, API-native JSON mode, forced function calling, and multi-turn tool orchestration, we observe a consistent principle: performance recovers whenever unconstrained reasoning precedes structured submission, regardless of the specific mechanism.

This finding is moderated by capacity margin: models with spare capacity absorb format constraints without degradation, while capacity-limited models suffer large penalties. The penalty operates through multiple channels---token-budget exhaustion for some models, reasoning-allocation interference for others---but the common trigger is simultaneous reasoning and serialization.

These findings have immediate practical implications for structured-output APIs and agentic workflows. API designs that natively support reasoning before structured submission (e.g., Claude's text-plus-tool response format) largely eliminate the format tax. Forced function calling without a reasoning phase can be catastrophic. For systems using instruction-prompted schemas, lightweight schemas impose far less tax than verbose ones, and delayed-structure prompting recovers most of the loss.

Future work should extend this investigation to non-mathematical reasoning tasks, constrained decoding methods, and open-weight architectures. Our results suggest a non-monotonic regime structure---format effects are small at ceiling, largest near the capability boundary, and collapse at the floor---which future work could formalize as a capacity-margin regime model and test for predictive power across models and tasks.

\section*{Acknowledgments}

We acknowledge the use of GPT-5.5 strictly for English language editing, proofreading, and improving the readability of the manuscript. The authors take full responsibility for the scientific content and the accuracy of the final text.

\bibliography{references}

\begin{thebibliography}{17}
\providecommand{\natexlab}[1]{#1}

\bibitem[{Banerjee et~al.(2025)Banerjee, Suresh, Ugare, Misailovic, and
  Singh}]{crane2025}
Banerjee, D.; Suresh, T.; Ugare, S.; Misailovic, S.; and Singh, G. 2025.
\newblock {CRANE}: Reasoning with Constrained {LLM} Generation.
\newblock In \emph{Proceedings of the International Conference on Machine
  Learning}.

\bibitem[{Cobbe et~al.(2021)Cobbe, Kosaraju, Bavarian, Chen, Jun, Kaiser,
  Plappert, Tworek, Hilton, Nakano, Hesse, and Schulman}]{cobbe2021training}
Cobbe, K.; Kosaraju, V.; Bavarian, M.; Chen, M.; Jun, H.; Kaiser, L.; Plappert,
  M.; Tworek, J.; Hilton, J.; Nakano, R.; Hesse, C.; and Schulman, J. 2021.
\newblock Training Verifiers to Solve Math Word Problems.
\newblock \emph{arXiv preprint arXiv:2110.14168}.

\bibitem[{Hendrycks et~al.(2021)Hendrycks, Burns, Kadavath, Arora, Basart,
  Tang, Song, and Steinhardt}]{hendrycks2021measuring}
Hendrycks, D.; Burns, C.; Kadavath, S.; Arora, A.; Basart, S.; Tang, E.; Song,
  D.; and Steinhardt, J. 2021.
\newblock Measuring Mathematical Problem Solving With the {MATH} Dataset.
\newblock In \emph{NeurIPS Datasets and Benchmarks}.

\bibitem[{Kaplan et~al.(2020)Kaplan, McCandlish, Henighan, Brown, Chess, Child,
  Gray, Radford, Wu, and Amodei}]{kaplan2020scaling}
Kaplan, J.; McCandlish, S.; Henighan, T.; Brown, T.~B.; Chess, B.; Child, R.;
  Gray, S.; Radford, A.; Wu, J.; and Amodei, D. 2020.
\newblock Scaling Laws for Neural Language Models.
\newblock \emph{arXiv preprint arXiv:2001.08361}.

\bibitem[{Lee, D'Antoni, and Berg-Kirkpatrick(2026)}]{lee2026format}
Lee, I.~Y.; D'Antoni, L.; and Berg-Kirkpatrick, T. 2026.
\newblock The Format Tax.
\newblock \emph{arXiv preprint arXiv:2604.03616}.

\bibitem[{Long et~al.(2025)Long, Ngoc, Sim, Dao, Joty, Kawaguchi, Chen, and
  Kan}]{long2025biased}
Long, D.~X.; Ngoc, H.~N.; Sim, T.; Dao, H.; Joty, S.; Kawaguchi, K.; Chen,
  N.~F.; and Kan, M.-Y. 2025.
\newblock {LLMs} Are Biased Towards Output Formats! Systematically Evaluating
  and Mitigating Output Format Bias of {LLMs}.
\newblock In \emph{Proceedings of the 2025 Conference of the North American
  Chapter of the Association for Computational Linguistics}.

\bibitem[{Ray(2026)}]{constrainttax2025}
Ray, J. 2026.
\newblock The Constraint Tax: Measuring Validity-Correctness Tradeoffs in
  Structured Outputs for Small Language Models.
\newblock \emph{arXiv preprint arXiv:2605.26128}.

\bibitem[{Suzgun et~al.(2023)Suzgun, Scales, Sch{\"a}rli, Gehrmann, Tay, Chung,
  Chowdhery, Le, Chi, Zhou, and Wei}]{suzgun2023challenging}
Suzgun, M.; Scales, N.; Sch{\"a}rli, N.; Gehrmann, S.; Tay, Y.; Chung, H.~W.;
  Chowdhery, A.; Le, Q.; Chi, E.; Zhou, D.; and Wei, J. 2023.
\newblock Challenging {BIG-Bench} Tasks and Whether Chain-of-Thought Can Solve
  Them.
\newblock In \emph{Findings of the Association for Computational Linguistics:
  ACL 2023}.

\bibitem[{Tam et~al.(2024)Tam, Wu, Tsai, Lin, Lee, and Chen}]{tam2024let}
Tam, Z.~R.; Wu, C.-K.; Tsai, Y.-L.; Lin, C.-Y.; Lee, H.-y.; and Chen, Y.-N.
  2024.
\newblock Let Me Speak Freely? A Study on the Impact of Format Restrictions on
  Performance of Large Language Models.
\newblock In \emph{Proceedings of the 2024 Conference on Empirical Methods in
  Natural Language Processing: Industry Track}.

\bibitem[{Wang et~al.(2024)Wang, Ma, Zhang, Ni, Chandra, Guo, Ren, Arulraj, He,
  Jiang et~al.}]{wang2024mmlu}
Wang, Y.; Ma, X.; Zhang, G.; Ni, Y.; Chandra, A.; Guo, S.; Ren, W.; Arulraj,
  A.; He, X.; Jiang, Z.; et~al. 2024.
\newblock {MMLU-Pro}: A More Robust and Challenging Multi-Task Language
  Understanding Benchmark.
\newblock \emph{Advances in Neural Information Processing Systems}, 37.

\bibitem[{Wei et~al.(2022{\natexlab{a}})Wei, Tay, Bommasani, Raffel, Zoph,
  Borgeaud, Yogatama, Bosma, Zhou, Metzler et~al.}]{wei2022emergent}
Wei, J.; Tay, Y.; Bommasani, R.; Raffel, C.; Zoph, B.; Borgeaud, S.; Yogatama,
  D.; Bosma, M.; Zhou, D.; Metzler, D.; et~al. 2022{\natexlab{a}}.
\newblock Emergent Abilities of Large Language Models.
\newblock \emph{Transactions on Machine Learning Research}.

\bibitem[{Wei et~al.(2022{\natexlab{b}})Wei, Wang, Schuurmans, Bosma, Ichter,
  Xia, Chi, Le, and Zhou}]{wei2022chain}
Wei, J.; Wang, X.; Schuurmans, D.; Bosma, M.; Ichter, B.; Xia, F.; Chi, E.; Le,
  Q.; and Zhou, D. 2022{\natexlab{b}}.
\newblock Chain-of-Thought Prompting Elicits Reasoning in Large Language
  Models.
\newblock \emph{Advances in Neural Information Processing Systems}, 35.

\bibitem[{Yang et~al.(2025)Yang, Yu, Cui, and Wang}]{reasonflux2025}
Yang, L.; Yu, Z.; Cui, B.; and Wang, M. 2025.
\newblock {ReasonFlux}: Hierarchical {LLM} Reasoning via Scaling Thought
  Templates.
\newblock \emph{arXiv preprint arXiv:2502.06772}.

\bibitem[{Yao et~al.(2023)Yao, Yu, Zhao, Shafran, Griffiths, Cao, and
  Narasimhan}]{yao2023tree}
Yao, S.; Yu, D.; Zhao, J.; Shafran, I.; Griffiths, T.~L.; Cao, Y.; and
  Narasimhan, K. 2023.
\newblock Tree of Thoughts: Deliberate Problem Solving with Large Language
  Models.
\newblock In \emph{Advances in Neural Information Processing Systems},
  volume~36.

\bibitem[{Yuan et~al.(2026)Yuan, Zhao, Zhang, Luo, and Ma}]{yuan2025causal}
Yuan, H.; Zhao, Y.; Zhang, L.; Luo, W.; and Ma, Z. 2026.
\newblock Quantifying the Impact of Structured Output Format on Large Language
  Models through Causal Inference.
\newblock \emph{Findings of the European Chapter of the Association for
  Computational Linguistics}.

\bibitem[{Zhou(2026)}]{snowballing2026}
Zhou, H. 2026.
\newblock From Hallucination to Structure Snowballing: The Alignment Tax of
  Constrained Decoding in {LLM} Reflection.
\newblock \emph{arXiv preprint arXiv:2604.06066}.

\bibitem[{Zhou et~al.(2024)Zhou, Pujara, Ren, Chen, Cheng, Le, Chi, Zhou,
  Mishra, and Zheng}]{zhou2024self}
Zhou, P.; Pujara, J.; Ren, X.; Chen, X.; Cheng, H.-T.; Le, Q.~V.; Chi, E.~H.;
  Zhou, D.; Mishra, S.; and Zheng, H.~S. 2024.
\newblock Self-Discover: Large Language Models Self-Compose Reasoning
  Structures.
\newblock In \emph{Advances in Neural Information Processing Systems},
  volume~37.

\end{thebibliography}

\appendix
\section{Full Prompt Templates}
\label{sec:prompts}

We list the complete system prompts for all conditions. The user message is always the problem text verbatim, with no additional framing.

\subsection{Main Experiment Conditions}

\paragraph{Chain-of-thought (CoT).} 82 characters.
\begin{quote}
\small\ttfamily
Solve the math problem. Think step by step, then give your final numerical answer.
\end{quote}

\paragraph{JSON-structured (JSON).} 1,774 characters.
\begin{quote}
\small\ttfamily
Solve the math problem. You must respond in the following JSON format with all fields filled in completely:\par
\medskip
\{\par
\quad"problem\_understanding": \{\par
\quad\quad"restatement": "restate the problem in your own words",\par
\quad\quad"question\_type": "classify as: arithmetic, algebra, geometry, combinatorics, or word\_problem",\par
\quad\quad"difficulty\_estimate": "easy, medium, or hard",\par
\quad\quad"key\_relationships": "describe the mathematical relationships between quantities"\par
\quad\},\par
\quad"known\_values": [\par
\quad\quad\{"name": "descriptive name", "value": "<number>", "unit": "unit", "role": "how used"\}\par
\quad],\par
\quad"unknown\_values": [\par
\quad\quad\{"name": "what we need to find", "unit": "expected unit", "constraints": "any constraints"\}\par
\quad],\par
\quad"approach": \{\par
\quad\quad"strategy": "describe your overall solution strategy",\par
\quad\quad"operations\_needed": ["list each operation required in order"],\par
\quad\quad"potential\_pitfalls": "describe common mistakes to avoid"\par
\quad\},\par
\quad"step\_by\_step": [\par
\quad\quad\{"step\_number": 1, "description": "what this step calculates and why", "operation": "the expression", "inputs": ["values used"], "result": "<number>", "unit": "unit", "verification": "sanity check"\}\par
\quad],\par
\quad"verification": \{\par
\quad\quad"sanity\_check": "does the answer make sense?",\par
\quad\quad"alternative\_method": "another way to verify",\par
\quad\quad"edge\_cases\_considered": "any special cases checked"\par
\quad\},\par
\quad"final\_answer": <number>\par
\}\par
\medskip
You MUST output valid JSON. Do not include any text outside the JSON object. Every field must be filled in.
\end{quote}

\paragraph{XML-structured (XML).} 1,625 characters. Same fields as JSON but using XML tags:
\begin{quote}
\small\ttfamily
Solve the math problem. You must respond in the following XML format with all tags filled in completely:\par
\medskip
<solution>\par
\quad<problem\_understanding>\par
\quad\quad<restatement>...</restatement>\par
\quad\quad<question\_type>...</question\_type>\par
\quad\quad<difficulty\_estimate>...</difficulty\_estimate>\par
\quad\quad<key\_relationships>...</key\_relationships>\par
\quad</problem\_understanding>\par
\quad<known\_values>\par
\quad\quad<value name="..." unit="..." role="...">number</value>\par
\quad</known\_values>\par
\quad<unknown\_values>\par
\quad\quad<unknown name="..." unit="..." constraints="..."/>\par
\quad</unknown\_values>\par
\quad<approach>\par
\quad\quad<strategy>...</strategy>\par
\quad\quad<operations>...</operations>\par
\quad\quad<potential\_pitfalls>...</potential\_pitfalls>\par
\quad</approach>\par
\quad<steps>\par
\quad\quad<step number="1">\par
\quad\quad\quad<description>...</description>\par
\quad\quad\quad<operation>...</operation>\par
\quad\quad\quad<result unit="...">number</result>\par
\quad\quad\quad<verification>...</verification>\par
\quad\quad</step>\par
\quad</steps>\par
\quad<verification>\par
\quad\quad<sanity\_check>...</sanity\_check>\par
\quad\quad<alternative\_method>...</alternative\_method>\par
\quad</verification>\par
\quad<final\_answer>number</final\_answer>\par
</solution>\par
\medskip
You MUST output valid XML. Do not include any text outside the XML. Every tag must be filled in.
\end{quote}

\paragraph{Detailed prose (DP).} 1,280 characters.
\begin{quote}
\small\ttfamily
Solve the math problem. Structure your response with the following clearly labeled sections:\par
\medskip
Problem Understanding: Restate the problem in your own words. Classify it by type (arithmetic, algebra, geometry, combinatorics, or word problem). Estimate its difficulty level. Describe the key mathematical relationships between the quantities mentioned.\par
\medskip
Known Values: List every given numerical value. For each value, state its name, the number, its unit of measurement, and how it is used in the problem. Be exhaustive.\par
\medskip
Unknown Values: Identify what we need to find. State the expected unit of the answer and any constraints.\par
\medskip
Approach: Before calculating, describe your overall strategy. List each mathematical operation you will need to perform, in order. Identify potential pitfalls or common mistakes for this type of problem.\par
\medskip
Step-by-Step Solution: Show each calculation step in detail. For each step, describe what is being calculated and why, write the mathematical expression, list the input values, state the numerical result with its unit, and verify that the intermediate result is reasonable.\par
\medskip
Verification: Perform a sanity check. Describe an alternative method to verify. Note any edge cases.\par
\medskip
Final Answer: State your final numerical answer clearly on its own line.
\end{quote}

\subsection{Delayed Structure Ablation}

\paragraph{Delayed JSON.} 522 characters.
\begin{quote}
\small\ttfamily
Solve the math problem in two phases.\par
\medskip
Phase 1 --- THINK: Solve the problem using whatever reasoning approach works best. Write your full solution in natural language. Do not worry about formatting.\par
\medskip
Phase 2 --- FORMAT: After you have your answer, reformat your solution into this JSON structure:\par
\medskip
\{"problem\_understanding": "your restatement", "step\_by\_step": [\{"step": 1, "operation": "description", "result": "number"\}], "final\_answer": <number>\}\par
\medskip
Clearly separate Phase 1 and Phase 2 with "--- Phase 2 ---".
\end{quote}

\subsection{Schema Complexity Gradient}

\paragraph{Minimal.} 73 characters.
\begin{quote}
\small\ttfamily
Solve the math problem. Respond with ONLY this JSON: \{"answer": <number>\}
\end{quote}

\paragraph{Light.} 117 characters.
\begin{quote}
\small\ttfamily
Solve the math problem. Respond in this JSON format:\par
\{"reasoning": "your step-by-step reasoning", "answer": <number>\}
\end{quote}

\paragraph{Medium.} 177 characters.
\begin{quote}
\small\ttfamily
Solve the math problem. Respond in this JSON format:\par
\{"approach": "describe your approach", "steps": [\{"step": 1, "calculation": "...", "result": <number>\}], "answer": <number>\}
\end{quote}

\paragraph{Heavy.} Same as the full JSON-structured prompt above (1,774 characters).

\subsection{BBH and MMLU-Pro Conditions}

For BBH and MMLU-Pro, prompts follow the same structure but adapted for multiple-choice answers. CoT ends with ``give your final answer as a single letter (A--J).'' JSON schemas replace \texttt{final\_answer: <number>} with \texttt{final\_answer: "(X)"} and add \texttt{option\_analysis} fields for MMLU-Pro. Full prompts are available in our code repository.

\section{Additional Statistical Tests}
\label{sec:stats}

All pairwise accuracy comparisons use McNemar's exact binomial test on discordant pairs (items correct under one condition but not the other). Wilcoxon signed-rank tests are used for paired token-count comparisons (reasoning efficiency). The logistic regression interaction test uses \texttt{statsmodels} with clustered standard errors by model$\times$benchmark cell. Single-run results are used for GSM8K, BBH, MMLU-Pro, and frontier probes; replicability intervals (1 deterministic + 2 stochastic reruns) are reported for MATH-Hard, the token-budget ablation, the delayed-structure ablation, and the schema gradient.

\begin{table}[h]
\centering
\small
\begin{tabular}{@{}llcc@{}}
\toprule
\textbf{Comparison} & \textbf{Model} & $\Delta$ & $p$ \\
\midrule
\multicolumn{4}{l}{\emph{MATH-Hard: JSON vs.\ CoT (3-run means)}} \\
& Sonnet 4.6  & $-$0.6 & 0.25 \\
& Haiku 4.5   & $-$36.2 & $<$0.0001 \\
& GPT-4o-mini & $-$28.0 & 0.0002 \\
\midrule
\multicolumn{4}{l}{\emph{MATH-Hard: JSON vs.\ DP (format-specific, 3-run means)}} \\
& Haiku 4.5   & $-$28.2 & $<$0.0001 \\
& GPT-4o-mini & $-$19.3 & 0.0011 \\
\midrule
\multicolumn{4}{l}{\emph{Delayed: Delayed vs.\ Simultaneous (3-run means)}} \\
& Haiku 4.5   & +27.7 & $<$0.0001 \\
& GPT-4o      & +16.3 & 0.0165 \\
\midrule
\multicolumn{4}{l}{\emph{Delayed: Delayed vs.\ CoT (3-run means)}} \\
& Haiku 4.5   & $-$7.0 & 0.4173 \\
& GPT-4o      & +5.3 & 0.5697 \\
& GPT-4o-mini & $-$2.9 & 0.9124 \\
\midrule
\multicolumn{4}{l}{\emph{Schema Gradient: McNemar (light $\to$ heavy)}} \\
& Haiku 4.5   & $-$37.0 & $<$0.0001 \\
& GPT-4o-mini & $-$24.0 & 0.0001 \\
& Sonnet 4.6  & +2.0 & 0.7539 \\
\midrule
\multicolumn{4}{l}{\emph{Interaction: is\_json $\times$ capability$_{\text{cross}}$ (logistic + benchmark FE)}} \\
& Item-level ($N$=4772) & $z$=+4.62 & $3.9 \times 10^{-6}$ \\
& Clustered by cell ($N$=16) & $z$=+2.90 & $3.8 \times 10^{-3}$ \\
& Cell-level Spearman ($N$=16) & $r_s$=+0.50 & 0.047 \\
\bottomrule
\end{tabular}
\caption{Complete statistical tests. The format$\times$capability interaction is significant on hard tasks using cross-validated capability (clustered: $p = 3.8 \times 10^{-3}$; cell-level: $p = 0.047$) but not on easy tasks.}
\label{tab:stats}
\end{table}

\end{document}